\documentclass[letterpaper, 10 pt, conference]{ieeeconf}  

\IEEEoverridecommandlockouts                              

\usepackage{graphics} 
\usepackage{epsfig} 
\usepackage{mathptmx} 
\usepackage{times} 
\usepackage{amsmath} 
\usepackage{amssymb}  

\usepackage[utf8]{inputenc}
\usepackage[noend]{algpseudocode}
\usepackage[utf8]{inputenc}
\usepackage{xcolor}
\usepackage{graphicx}
\usepackage{hyperref}
\usepackage{booktabs}
\usepackage{color}
\usepackage{array}
\usepackage{booktabs}
\usepackage{multirow}
\usepackage{geometry}
\usepackage{hyperref}
\usepackage[space]{cite}
\usepackage[linesnumbered,boxed,ruled,commentsnumbered]{algorithm2e}
\usepackage[utf8]{inputenc}
\usepackage[noend]{algpseudocode}
\usepackage{amsmath}
\usepackage{amssymb}    
\usepackage[utf8]{inputenc}
\usepackage{xcolor}
\usepackage{graphicx}
\usepackage{hyperref}
\usepackage{booktabs}
\usepackage{color}
\usepackage{array}
\usepackage{booktabs}
\usepackage{multirow}
\usepackage{geometry}
\usepackage{hyperref}
\usepackage[space]{cite}
\usepackage{mathtools}
\usepackage{threeparttable} 
\usepackage{makecell} 
\usepackage{wasysym}
\usepackage{lipsum}
\usepackage{tabularx} 
\usepackage{makecell} 
\DeclareGraphicsExtensions{.pdf,.png,.jpg,.eps}
\title{\LARGE \bf
CM-LIUW-Odometry: Robust and High-Precision LiDAR-Inertial-UWB-Wheel Odometry for Extreme Degradation Coal Mine Tunnels
}

\begin{document}

\author{Kun Hu$^{1}$,
    Menggang Li$^{1\dag}$,
    Zhiwen Jin$^{1}$,
    Chaoquan Tang$^{1}$,
    Eryi Hu$^{2}$ and Gongbo Zhou$^{1}$
    \thanks{Research supported by National Natural Science Foundation of China (grant number: 52304183, 52274159), the China Postdoctoral Science Foundation (2022M723392), The Jiangsu Provincial Key Research and Development (R\&D) Plan Projects (BE2023008-4). The Project Funds of the Priority Academic Program Development of Jiangsu Higher Education Institutions (PAPD).}
    \thanks{$^{1}$ K. Hu, M. Li, Z. Jin, C. Tang and G. Zhou are with School of Mechatronic Engineering, China University of Mining and Technology, Xuzhou, 221116, China.
            {\tt\footnotesize $\{$ts22050017a31, sallylmg, ts24050202p31, tangchaoquan, gbzhou$\}$@cumt.edu.cn}}
    \thanks{$^{2}$ E. Hu is with Information Institute, Ministry of Emergency Management of the People's Republic of China, Beijing 100029, China.
            {\tt\footnotesize horyhu@126.com}}
    \thanks{$^{\dag}$ Menggang Li is the corresponding author.}
}

\markboth{Journal of \LaTeX\ Class Files,~Vol.~6, No.~1, January~2007}%
{Shell \MakeLowercase{\textit{et al.}}: Bare Demo of IEEEtran.cls for Journals}

\maketitle
\thispagestyle{empty}
\pagestyle{empty}

\begin{abstract}
Simultaneous Localization and Mapping (SLAM) in large-scale, complex, and GPS-denied underground coal mine environments presents significant challenges. Sensors must contend with abnormal operating conditions: GPS unavailability impedes scene reconstruction and absolute geographic referencing, uneven or slippery terrain degrades wheel odometer accuracy, and long, feature-poor tunnels reduce LiDAR effectiveness. To address these issues, we propose CoalMine-LiDAR-IMU-UWB-Wheel-Odometry (CM-LIUW-Odometry), a multimodal SLAM framework based on the Iterated Error-State Kalman Filter (IESKF). First, LiDAR-inertial odometry is tightly fused with UWB absolute positioning constraints to align the SLAM system with a global coordinate. Next, wheel odometer is integrated through tight coupling, enhanced by nonholonomic constraints (NHC) and vehicle lever arm compensation, to address performance degradation in areas beyond UWB measurement range. Finally, an adaptive motion mode switching mechanism dynamically adjusts the robot’s motion mode based on UWB measurement range and environmental degradation levels. Experimental results validate that our method achieves superior accuracy and robustness in real-world underground coal mine scenarios, outperforming state-of-the-art approaches. We open source our code of this work on Github\footnote[3]{\url{https://github.com/KJ-Falloutlast/CM-LIUW-Odometry}} to benefit the robotics community.
\end{abstract}

\section{INTRODUCTION}
\textbf{Simultaneous Localization and Mapping (SLAM)} has rapidly developed in recent years, aiming to enable real-time robot localization and environmental mapping. Due to its significant potential in enhancing autonomous systems, SLAM technology has attracted increasing attention in industrial applications. However, implementing SLAM for underground coal mine robots presents greater challenges compared to ground-based mobile robots:
\begin{itemize}

    \item \textbf{Challenge 1:} Subterranean environments are characterized by poor illumination and complex conditions, including dust, humidity, and post-disaster visibility degradation. Slippery and uneven road surfaces further compromise the performance of conventional sensors (e.g., LiDAR and cameras).  High signal-to-noise ratio sensors and effective information processing methods are essential for ensuring long-term robustness.  

    \item \textbf{Challenge 2:} The deployment range of UWB positioning systems in underground coal mine tunnels is limited, and their cost is high. Repetitive features and long tunnels can lead to degradation of LiDAR-Inertial Odometry (LIO). There is an urgent need to integrate additional sensors to compensate for positioning information beyond the UWB measurement range. 

	\item \textbf{Challenge 3:} Traditional degradation detection methods rely on threshold-based Hessian matrix eigenvalue analysis or parameter tuning of robot, sensor, and environmental configurations. These approaches lack adaptive adjustment of motion modes based on environmental degradation levels and demonstrate limited generalizability across heterogeneous scenarios.

\end{itemize}

To address these challenges, we propose a tightly coupled CoalMine-LiDAR-IMU-UWB-Wheel-Odometry (CM-LIUW-Odometry) multimodal SLAM method based on the Iterated Error-State Kalman Filter (IESKF) framework. The main contributions of this work are as follows:

\begin{itemize}
	\item \textbf{Contribution 1:} Through tight coupling of LiDAR-inertial odometry with UWB absolute positioning constraints. We achieve alignment of the SLAM system with global coordinates, while establishing a UWB positioning system to ensure precise localization within the UWB measurement range.
	
	\item \textbf{Contribution 2:} To mitigate the limited measurement range and high deployment costs of UWB systems in coal mines, we introduce nonholonomic constraints (NHC) and vehicle lever arm compensation. We integrate wheel odometer into the fusion framework through tight coupling to reduce SLAM system degradation beyond the UWB measurement range.

	\item \textbf{Contribution 3:} A novel degradation detection and adaptive motion mode switching mechanism is proposed. This mechanism evaluates degradation direction and severity through degradation detection, then determines whether to implement wheel odometer constraints to enhance SLAM system robustness.

    \item \textbf{Contribution 4:} Extensive experiments validate the performance of the proposed method in real-world coal mine degradation scenarios. Comprehensive evaluations and comparisons with state-of-the-art methods demonstrate its superior performance and reliability in challenging environments.

\end{itemize}

The paper is organized as follows: Chapter \ref{chap2-Related_work} discusses related work; Chapters \ref{chap4-system_overview} and \ref{chap5-Methodology} present the complete system framework and detailed implementation of key components, respectively; Chapter \ref{chap6-experiment} describes experimental validation; Chapter \ref{chap7-conclusion} presents the conclusion.

\section{Related Work}\label{chap2-Related_work}

Simultaneous Localization and Mapping (SLAM) in complex underground coal mine environments presents significant challenges. While LiDAR is capable of capturing detailed information and performing long-range measurements, it is prone to failure in unstructured environments, such as long, straight tunnels. Visual methods, on the other hand, are limited by lighting variations and motion blur. Ultra-Wideband (UWB), known for its resistance to multipath interference, has emerged as a potential solution for coal mine localization. However, its deployment range and high cost restrict its practical application. Wheel odometer can provide more stable motion predictions than IMUs in LiDAR-degraded environments, but it suffers from large cumulative errors due to terrain roughness. This work is closely related to methods in LiDAR-Inertial Odometry (LIO), UWB-Inertial Odometry (UIO), Wheel-Inertial Odometry (WIO), and degradation detection.

LiDAR-Inertial Odometry (LIO) can be categorized into loosely-coupled and tightly-coupled approaches, with tightly-coupled methods typically offering superior accuracy and robustness. For example, LIO-SAM \cite{shan2020lio} pioneered a factor graph framework for fusing heterogeneous measurements, while FAST-LIO2 \cite{xu2022fast} enhanced mapping precision through raw point registration within an efficient tightly-coupled iterated Kalman filter framework. DLIO \cite{chen2023direct} achieved precise motion correction through a coarse-to-fine strategy, and Adaptive-LIO \cite{zhao2024adaptive} improved localization accuracy via adaptive segmentation and multi-resolution mapping.

UWB-Inertial Odometry (UIO) methodologies include both filtering-based and optimization-based approaches. Li et al. \cite{li2020uwb} achieved high-precision localization by fusing UWB and IMU measurements using an Extended Kalman Filter (EKF). In our prior work \cite{li2023multimodal}, we developed an incremental factor graph optimization framework to integrate LiDAR-inertial odometry with UWB anchor-based absolute constraints. Other studies, such as \cite{wang2017ultra} and \cite{cao2021vir}, demonstrated UWB-visual SLAM fusion, which effectively reduces drift but is still constrained by UWB deployment limitations, failing to meet large-scale positioning requirements.

Wheel-Inertial Odometry (WIO) can be divided into kinematic model-based methods and integration methods. Several modeling approaches have been proposed \cite{anousaki2004dead, mandow2007experimental}. Mandow et al. \cite{mandow2007experimental} proposed an extended differential drive model to improve the motion modeling of skid-steering robots. ACK-MSCKF \cite{ma2019ack} incorporated pre-integration of wheel odometer measurements for forward velocity and angular rate estimation, although it relies on strict flat-ground assumptions. Liu et al. \cite{liu2019visual} enhanced accuracy through pre-integration of wheel encoder and gyroscope measurements, but pre-integration errors become significant under large variations in vehicle speed.

For degradation detection, Zhang et al. \cite{zhang2016degeneracy} identified odometry degradation by analyzing the eigenvalues of LiDAR's Hessian matrix. X-ICP \cite{tuna2023x} combined local localizability detection with optimization, improving scan matching accuracy in degraded environments. LVI-SAM \cite{shan2021lvi} combined LIO with VIO to effectively address LiDAR degradation, although it remains prone to failures in visual SLAM. FAST-LIVO \cite{zheng2022fast} and FAST-LIVO2 \cite{zheng2024fast} improved system robustness in LiDAR- or visual-degraded environments. Our proposed CM-LIUW-Odometry performs degradation detection using covariance principal component analysis, enabling adaptive switching between LIU and LIW modes when degradation is detected. This approach addresses degradation issues in coal mine tunnels and enhances localization robustness.

\section{Preliminary}\label{chap3-Preliminary}
\subsection{Notations and Definitions}
\begin{table}[h]
    \centering
    \caption{Definitions of Important Variables}
    \label{tbl1:symbols}
    \begin{tabular}{lll}
        \toprule
        Notations  & Meaning \\
        \midrule
        $\mathbf{x}, \hat{\mathbf{x}}, \mathbf{\overline{x}}$ & The ground-truth, predicted and updated estimation of $\mathbf{x}$.\\
        $\delta{\mathbf{x}}$ & Error state.\\
        $(^I\mathbf{R}_L, ^I\mathbf{p}_L)$ & The extrinsic of the LiDAR frame w.r.t. the IMU frame. \\
        $(^I\mathbf{R}_U, ^I\mathbf{p}_U)$ & The extrinsic of the UWB frame w.r.t. the IMU frame.\\
        $(^I\mathbf{R}_W, ^I\mathbf{p}_W)$ & The extrinsic of the Wheel frame w.r.t. the IMU frame. \\
        \bottomrule
    \end{tabular}
\end{table}
In our system, we assume that the time offsets between the four sensors (LiDAR, IMU, Wheel, and UWB) are known and can either be pre-calibrated or synchronized. We adopt the IMU coordinate system (denoted as $ I $) as the body coordinate system and define the center of the coordinate system, where the total station is located, as the global coordinate system (denoted as $ G $). Furthermore, the four sensors are firmly integrated, with LiDAR and IMU having undergone hardware time synchronization. The extrinsic parameters, as defined in Table \ref{tbl1:symbols}, have been pre-calibrated. The discrete state transition model at the $i$-th IMU measurement is then given by:
\begin{equation}
    \mathbf{x}_{i+1}=\mathbf{x}_i \boxplus\left(\Delta t \mathbf{f}\left(\mathbf{x}_i, \mathbf{u}_i, \mathbf{w}_i\right)\right),
    \label{eq_1_state_propagate}
\end{equation}
where $ \boxplus/\boxminus $ is generalized addition and subtraction \cite{xu2022fast}, and $ \Delta t $ is the IMU sampling period. The state $ \mathbf{x} \in \mathcal{R}^{36} $ is defined as follows:
\begin{equation}
    \small
    \mathbf{x} \triangleq\left[\begin{array}{lllllllll}
    { }^G \mathbf{R}_I^T & { }^G \mathbf{p}_I^T & { }^G \mathbf{v}_I^T & \mathbf{b}_{\mathbf{g}}^T & \mathbf{b}_{\mathbf{a}}^T & { }^G \mathbf{g}^T & { }^I \mathbf{T}_L^T & { }^I \mathbf{T}_U^T & { }^I \mathbf{T}_W^T
    \end{array}\right]^T,
    \label{eq_2_state}
\end{equation}
where $ ^I\mathbf{T}_L=(^I\mathbf{R}_L, ^I\mathbf{p}_L) $,$ ^I\mathbf{T}_U=(^I\mathbf{R}_U, ^I\mathbf{p}_U) $ and $ ^I\mathbf{T}_W=(^I\mathbf{R}_W, ^I\mathbf{p}_W) $ are explained in Table \ref{tbl1:symbols}. The input $ \mathbf{u} $, process noise $ \mathbf{w} $, and function $ \mathbf{f} $ are defined in \cite{xu2022fast}, and due to space limitations, they are not elaborated here.

\subsection{Error-state Iterated Kalman Filter Update}
From the Forward Propagation \cite{xu2022fast}, the propagated state $ \hat{\mathbf{x}}_k $ and covariance $ \hat{\mathbf{P}}_k $ apply a prior distribution to $ \mathbf{x}_k $ as follows:
\begin{equation}
    \mathbf{x}_k \boxminus \hat{\mathbf{x}}_k \sim \mathcal{N}\left(\mathbf{0}, \hat{\mathbf{P}}_k\right), 
    \label{eq_3_prior_distribution}
\end{equation}
by combining the prior distribution in (\ref{eq_3_prior_distribution}), UWB position measurements $ \mathbf{z}_{U_p} $, UWB distance measurements $ \mathbf{z}_{U_r} $ (refer to Chapter \ref{chap5-2-uwb_constrains}), wheel odometer measurements $ \mathbf{z}_{W} $ (refer to Chapter \ref{chap5-3-wheel_constrains}), and LiDAR measurements $ \mathbf{z}_{L}^i $ (refer to Chapter \ref{chap5-4-lidar_constrains}), we obtain the Maximum A Posteriori (MAP) estimate of $ \delta \mathbf{x}_k$:
\begin{equation}
    \begin{aligned}
    \min _{\mathbf{\delta x}_k \in \mathcal{M}} \Bigg( & \left\|\mathbf{x}_k \boxminus \hat{\mathbf{x}}_k\right\|_{\hat{\mathbf{P}}_k}^2 
    + \sum_{i=1}^{N_{L}} \left\|\mathbf{r}_{L}\left(\mathbf{z}_{L}^i, \mathbf{x}_k\right)\right\|_{\mathbf{P}_{L}^i{}^{-1}}^2 
    + \sum_{i=1}^{N_{U_p}} \left\|\mathbf{r}_{U_p}\left(\mathbf{z}_{U_p}^i, \mathbf{x}_k\right)\right\|_{\mathbf{P}_{U_p}^i{}^{-1}}^2 \\
    & + \sum_{i=1}^{N_{U_r}} \left\|\mathbf{r}_{U_r}\left(\mathbf{z}_{U_r}^i, \mathbf{x}_k\right)\right\|_{\mathbf{P}_{U_r}^i{}^{-1}}^2 
    + \sum_{i=1}^{N_{W}} \left\|\mathbf{r}_{W}\left(\mathbf{z}_{W}^i, \mathbf{x}_k\right)\right\|_{\mathbf{P}_{W}^i{}^{-1}}^2 \Bigg),
    \end{aligned}
    \label{eq_4_constrains}
\end{equation}
where $ \|\mathbf{x}\|_{\mathbf{P}}^2 = \mathbf{x}^T \mathbf{\mathbf{P}}^{-1} \mathbf{x} $, and $ \mathbf{r}_{L}, \mathbf{r}_{U_p}, \mathbf{r}_{U_r}, \mathbf{r}_{W} $ are the LiDAR residuals, UWB position residuals, UWB distance residuals, and wheel odometer residuals, respectively. The covariance matrices corresponding to LiDAR measurements, UWB position measurements, UWB distance measurements, and wheel odometer measurements are denoted as $ \mathbf{P}_{L}^i$, $\mathbf{P}_{U_p}^i$, $\mathbf{P}_{U_r}^i$, and $\mathbf{P}_{W}^i $ . $ N_L, N_{U_p}, N_{U_r}, N_W $ represent the number of LiDAR, UWB, and wheel odometer observations obtained in the time interval from $t_{k-1}$ to $t_k$. The optimization in (\ref{eq_4_constrains}) is non-convex and can be iteratively solved using the Gauss-Newton method. This iterative optimization has been proven to be equivalent to the Iterated Kalman Filter \cite{bell1993iterated}.

\section{System Overview}\label{chap4-system_overview}
\begin{figure}
    \centering
    \includegraphics[width=1.0\columnwidth]{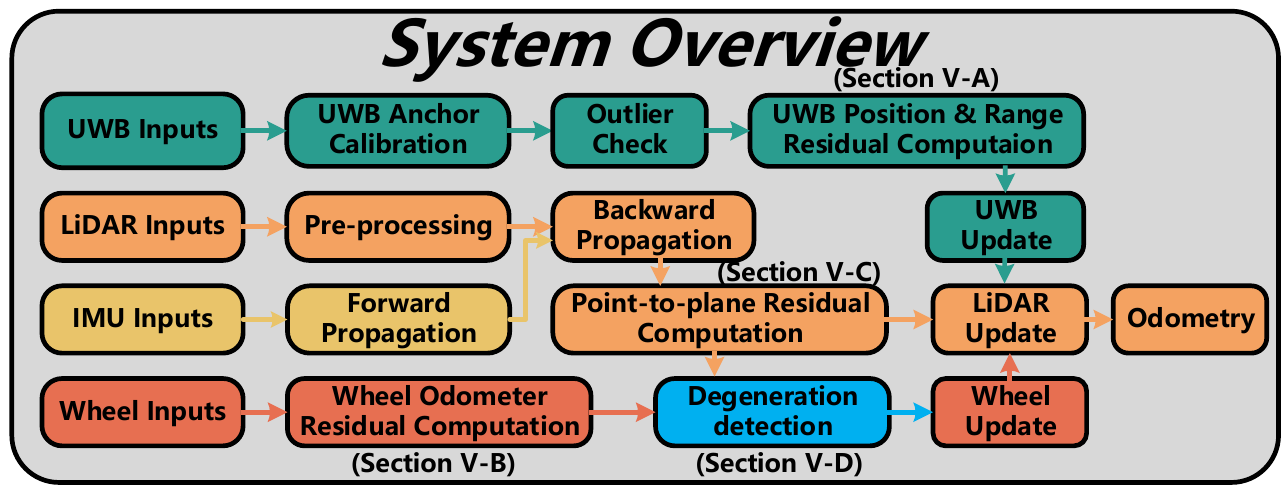}
    \caption{System overview.}
    \label{Fig1-System overview}
\end{figure}

The objective of this work is to estimate the 6-degree-of-freedom pose of the coal mine robot while simultaneously constructing a global map. The system framework, shown in Fig. \ref{Fig1-System overview}, consists of four primary modules: the UWB Constraint Module, the Wheel Odometer Constraint Module, the LiDAR Constraint Module, and the Degradation Detection and Adaptive Motion Mode Switching Module:

\begin{enumerate} 
    \item \textbf{UWB Constraint Module:} The positions of UWB anchors are calibrated using a total station, with the center of the total station's coordinate system serving as the global coordinate system, denoted as $ G $. The UWB data are synchronized with the timestamps of the other sensors. A preliminary check is performed on each UWB measurement (including distance and position observations) to eliminate poor-quality measurements. Finally, UWB distance and position constraints are constructed and tightly integrated into the system (refer to Chapter \ref{chap5-2-uwb_constrains}).
    
    \item \textbf{Wheel Odometer Constraint Module:} The wheel speed from the vehicle's wheel odometer is obtained, and the wheel data are synchronized with the LiDAR timestamps. Wheel odometer constraints are then constructed using nonholonomic constraints (NHC) and vehicle lever arm compensation. (refer to Chapter \ref{chap5-3-wheel_constrains}).

    \item \textbf{LiDAR Constraint Module:} The point cloud is first pre-processed for downsampling and feature extraction. The IMU is then used to undistort the point cloud, after which LiDAR point-plane constraints are constructed and tightly integrated into the system with the IMU (refer to Chapter \ref{chap5-4-lidar_constrains}).

    \item \textbf{Degradation Detection and Adaptive Motion Mode Switching Module:} The degree and direction of degradation are determined using principal component analysis of the covariance. Based on the robot's position relative to the UWB measurement range and the level of degradation, the UWB and wheel odometer constraints are selectively incorporated, enabling adaptive motion mode switching (refer to Chapter \ref{chap5-5-degradation_detection}).
\end{enumerate}

\section{Methodology}\label{chap5-Methodology}
\subsection{UWB Constraint Module}\label{chap5-2-uwb_constrains}
\subsubsection{UWB Position Constraint}
The UWB positioning system serves as a replacement for the GPS system in underground environments, providing positioning within the global coordinate system $ G $ established by the UWB anchor nodes. As demonstrated in our previous work, the extended Kalman filter (EKF) is employed to construct the UWB positioning system \cite{li2020uwb}. The system outputs global position information in the global coordinate system $ G $, with the UWB position measurement denoted as:
\begin{equation}
    \mathbf{z}_{U_p}= h(\mathbf{x}_k, \mathbf{n}_{U_p}) = ^G{}\mathbf{p}_U+\mathbf{n}_{U_p}=^G\mathbf{p}_I+^G\mathbf{R}_I ^I\mathbf{p}_U + \mathbf{n}_{U_p}.
    \label{eq_5_uwb_position}
\end{equation}

UWB position observation $ ^G\mathbf{p}_U $ depends on the pose of the IMU in the global coordinate system $ G $, $ ^G{}\mathbf{T}_I $, as well as the translation part of the extrinsic parameters between the UWB antenna and the IMU, $ ^I\mathbf{p}_U $. $ \mathbf{n}_{U_p} \sim \mathcal{N}(0, \sigma_{U_p}) $ is the UWB position measurement noise. The UWB measurement directly constrains the state to be optimized, $ \mathbf{x}_k $. The UWB position residual $ \mathbf{r}_{U_p} $ can be expressed as:
\begin{equation}
    \mathbf{r}_{U_p}  = ^G\hat{\mathbf{p}}_U-^G\mathbf{p}_I-^G\mathbf{R}_I ^I\mathbf{p}_U.
    \label{eq_6_uwb_position_residual}
\end{equation}

\subsubsection{UWB Distance Constraint}

For narrow mine tunnel applications, the cost of large-scale UWB node coverage is substantial, and deploying a large number of UWB anchor nodes is neither economical nor necessary. The system's position estimation can be independently achieved through LiDAR-Inertial Odometry, as the position estimation error in structurally rich tunnels is relatively small. However, in degraded tunnels with similar scene structures, where LiDAR scan matching lacks sufficient distinguishable features, the error increases rapidly. To address this, we propose deploying a small number of UWB anchor nodes in areas prone to degradation, providing distance constraints along the tunnel direction (i.e., the vehicle's motion direction) to mitigate the impact of degraded scenes on position estimation and improve accuracy. Analysis of the UWB position covariance components reveals that UWB anchor nodes, deployed in the global coordinate system $ G $ with known positions $ ^G\mathbf{p}_{f_i} $, can provide constraints along the direction between the anchor nodes and the mobile node, as demonstrated in our previous work \cite{li2020uwb}. UWB distance measurements can be described as:
\begin{equation}
    \mathbf{z}_{U_r}^i=h(\mathbf{x}_k, \mathbf{n}_{U_r}^i) =\sqrt{\left(^G\mathbf{p}_{f_i}-^G\mathbf{p}_U\right)^T\left(^G\mathbf{p}_{f_i}-^G\mathbf{p}_U\right)}+\mathbf{n}_{U_r^i},
    \label{eq_8_uwb_distance}
\end{equation}
where $ ^G\mathbf{p}_U $ is the position in the global coordinate system $ G $, $ ^G\mathbf{p}_{f_i} $ is the position of the $ i $-th UWB anchor node, and $ \mathbf{n}_{U_{r}}^i \sim \mathcal{N}(0, \sigma_{U_{r}}^i) $ is the measurement noise of the $ i $-th UWB anchor node. Here, $ i $ corresponds to the indices 0, 1, 2, 3, representing the distance measurements between the UWB anchor nodes 100, 101, 102, 103 and the UWB mobile node 104. The points in the UWB coordinate system are given by:
\begin{equation}
    ^G\mathbf{p}_U=^G\mathbf{p}_I + ^G\mathbf{R}_I ^I\mathbf{p}_U.
    \label{eq_9_uwb_point}
\end{equation}

The UWB distance residual $ \mathbf{r}_{U_r}^i $ can be expressed as:
\begin{equation}
    \mathbf{r}_{U_r}^i =\hat d_i-\sqrt{\left(^G\mathbf{p}_{f_i}-^G\mathbf{p}_U\right)^T\left(^G\mathbf{p}_{f_i}-^G\mathbf{p}_U\right)}=\hat d_i-d_i, 
        \label{eq_10_uwb_distance_residual}
\end{equation}
where $ \hat{d}_i $ is the distance measurement from the $ i $-th UWB anchor node to the UWB mobile node.

\subsection{Wheel Odometer Constraint Module}\label{chap5-3-wheel_constrains}

\begin{figure}[t]
    \centering
    \includegraphics[width=0.5\columnwidth]{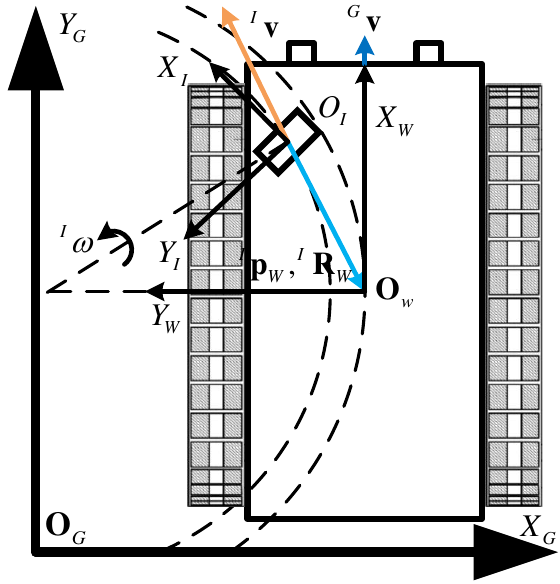}
    \caption{Lever arm compensation between the IMU and the wheel odometer coordinate system. When the vehicle turns, the IMU and the wheel odometer have different velocities, $^I\mathbf{v}$ and $^W\mathbf{v}$, respectively. Therefore, the wheel odometer velocity $^W\mathbf{v}$ needs to be transformed from the wheel odometer coordinate system $\mathbf{O}_W$ to the IMU coordinate system $\mathbf{O}_I$ using their extrinsic parameters $^I\mathbf{R}_W$ and $^I\mathbf{p}_W$ for fusion.}
    \label{Fig3-Wheel_odometry}
    \vspace{0cm}
\end{figure}

In underground coal mines, the deployment range of the UWB positioning system is limited and costly, particularly in degraded scenarios where repetitive features and long tunnels, sometimes extending for several kilometers, are common. The LIO system tends to degrade in such environments, and when the robot moves beyond the UWB measurement range, the distance and position measurements provided by the UWB system become inaccurate. This not only diminishes the localization accuracy of the SLAM system but may also lead to system failure. To address this challenge, we introduce nonholonomic constraints (NHC) and lever arm compensation, establishing a measurement model based on vehicle kinematics. By using the vehicle's velocity, we can directly constrain velocity errors without the need to integrate velocity into displacement. This method leverages wheel odometer constraints to assist the LIO system, mitigating LiDAR degradation issues in repetitive features and long straight tunnels.

As shown in Fig. \ref{Fig3-Wheel_odometry}, NHC can be modeled as $ {}^W\mathbf{v}_y = 0, {}^W\mathbf{v}_z = 0 $. In practice, the actual values $ {}^W\hat{\mathbf{v}}_y $ and $ {}^W\hat{\mathbf{v}}_z $ cannot be exactly zero, so we introduce Gaussian noise $ \sigma_{\mathbf{v}_{W}}^x, \sigma_{\mathbf{v}_{W}}^y, \sigma_{\mathbf{v}_{W}}^z $ to represent velocity noise. That is, $ {}^W\mathbf{v}_{i} = {}^W\hat{\mathbf{v}}_{i} - \mathbf{n}_{i} $, where $ \mathbf{n}_{i} \sim \mathcal{N}(0, \sigma_{\mathbf{v}_{W}}^{i}) $, and $ i $ represents $ x, y, z $. Additionally, NHC implies that the velocity at the origin of the vehicle coordinate system always matches the forward velocity of the vehicle. For a differential drive vehicle, the origin of the body coordinate system is located at the center of the vehicle (as shown in Fig. \ref{Fig3-Wheel_odometry}), so the origin of the body coordinate system, $ \mathbf{O}_W $, coincides with the vehicle's center. Based on NHC and lever arm compensation, we can derive the velocity relationship between the IMU coordinate system $ I $ and the wheel odometer coordinate system $ W $:
\begin{equation}
    ^I\mathbf{v}={}^I\mathbf{R}_W {}^W\mathbf{v}-{}^I\boldsymbol{\omega} \times {}^I\mathbf{p}_W,
    \label{eq_13_vehicle_speed}
\end{equation}
where $ {}^I\mathbf{v} $ and $ {}^W\mathbf{v} $ are the velocities in the IMU and wheel odometer coordinate systems, $ {}^I\mathbf{R}_W $ and $ {}^I\mathbf{p}_W $ are the rotation and translation extrinsics from the wheel odometer coordinate system to the IMU frame, and $ {}^I\boldsymbol{\omega} $ is the angular velocity in the IMU coordinate system. Since the velocity in the IMU coordinate system and the velocity in the global coordinate system are related by:
\begin{equation}
    {}^I\mathbf{v}={}^G\mathbf{R}_I^T {}^G\mathbf{v}.
    \label{eq_14_vehicle_imu_trans}
\end{equation}

By substituting (\ref{eq_14_vehicle_imu_trans}) into (\ref{eq_13_vehicle_speed}), we obtain the velocity observation in the wheel odometer coordinate system $ W $:
\begin{equation}
    \mathbf{z}_W = h(\mathbf{x}_k, \mathbf{n}_{W}) = {}^W\mathbf{v} = {}^I\mathbf{R}_W^T ({}^G\mathbf{R}_I^T {}^G\mathbf{v} + {}^I\boldsymbol{\omega} \times {}^I\mathbf{p}_W) + \mathbf{n}_W,
    \label{eq_15_vehicle_obs}
\end{equation}
where $ \mathbf{n}_W = [\sigma_{\mathbf{v}_{W}}^x, \sigma_{\mathbf{v}_{W}}^y, \sigma_{\mathbf{v}_{W}}^z]^T $, and we construct the wheel odometer velocity residual:
\begin{equation}
    \mathbf{r}_W = {}^W\hat{\mathbf{v}} - {}^I\mathbf{R}_W^T ({}^G\mathbf{R}_I^T {}^G\mathbf{v} + {}^I\boldsymbol{\omega} \times {}^I\mathbf{p}_W),
    \label{eq_16_vehicle_residual}
\end{equation}
where $ {}^W\hat{\mathbf{v}} $ is the velocity observation in the wheel odometer coordinate system.

\subsection{LiDAR Constraint Module}\label{chap5-4-lidar_constrains}
Consistent with the method outlined in \cite{xu2022fast}, if a LiDAR measurement is received at time $t_k$, we first perform backward propagation to compensate for motion distortion. The resulting points in the scan, $\{^L{}\mathbf{p}_j\}$, can be treated as being sampled simultaneously at $t_k$. We then construct an observation model based on the point-to-plane distance. If the true state (i.e., pose) $\mathbf{x}_k$ is used to transform the measurement $^L{}\mathbf{p}_j$, expressed in the LiDAR local coordinate system, to the global coordinate system $G$, the LiDAR residual $\mathbf{r}_{L}^j$ should be zero:
\begin{equation}
    \mathbf{0}=\mathbf{r}_L^j\left(\mathbf{x}_k, \mathbf{n}_j^L\right)=\mathbf{u}_j^T\left({ }^G \mathbf{T}_{I_k}{ }^I \mathbf{T}_L{ }^L \mathbf{p}_j-\mathbf{q}_j\right), 
    \label{eq_18_lidar_obs}
\end{equation}
where $\mathbf{n}_j^L$ represents the LiDAR measurement noise, and ${ }^L \mathbf{p}_j$ represents the coordinates of the point in the LiDAR frame. The vector $\mathbf{u}_j^T$ is the normal vector of the plane matched with the point ${ }^L \mathbf{p}_j$ in the map, and $\mathbf{q}_j$ is a point on that plane.

\subsection{Degradation Detection and Adaptive Motion Mode Switching Module}\label{chap5-5-degradation_detection}

\begin{figure}[t]
    \centering
    \includegraphics[width=1\columnwidth]{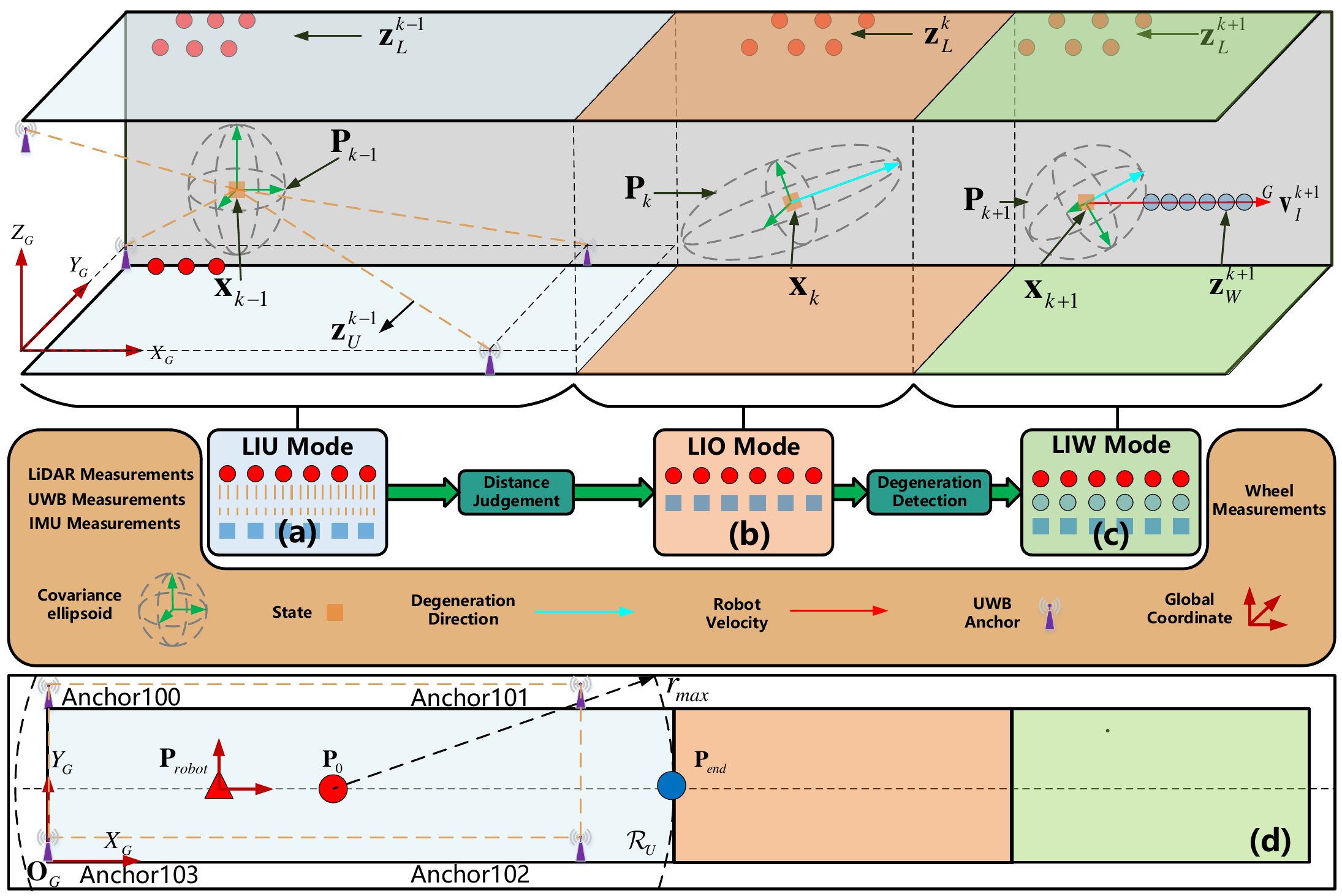}
    \caption{Schematic diagram of degradation detection and adaptive motion mode switching. (a) When the robot is within the  measurement range of the UWB positioning system, the system uses the LIU mode for localization, and the covariance ellipsoid is uniformly distributed; (b) After the robot exits the UWB measurement range, the system switches to LIO mode. Due to the loss of the X-axis constraint, the covariance ellipsoid expands significantly along the X-axis direction, leading to system degradation; (c) After detecting degradation in the X-axis direction, the system introduces the wheel odometer constraint $\mathbf{z}_{W}^{k+1}$ and switches to LIW mode, reducing the covariance ellipsoid in the degraded direction; (d) Shows the UWB and robot positions in the global coordinate system $G$ and the UWB measurement range $\mathcal{R}_U$.}
    \label{Fig4-Dengerate_detection}
    \vspace{0cm}
\end{figure}

In this section, we describe the working principle of the degradation detection and adaptive motion mode switching module, based on the information flow shown in Fig. \ref{Fig4-Dengerate_detection}. This module consists of two core components: degradation detection and adaptive motion mode switching. Its purpose is to assess the degree and direction of degradation through Singular Value Decomposition (SVD) of the covariance matrix and to switch to an appropriate motion mode based on range detection and the degradation degree. This approach enhances the robustness of the SLAM system in various environments.

\subsubsection{Degradation Detection}

In the SLAM system, the covariance matrix $\mathbf{P}$ reflects the uncertainty of the system state. During the state propagation process, the covariance matrix $\mathbf{P}$ is updated by sensor observations. We use Singular Value Decomposition (SVD) to decompose the rotational and translational components of the covariance matrix. By analyzing these components, we can identify the degree and direction of degradation in the system. Specifically, SVD helps pinpoint directions in the state space with high uncertainty, which serves as the foundation for degradation detection.

We extract the submatrix $\mathbf{B} \in \mathbb{R}^{6 \times 6}$, which includes rotational and translational components, from the covariance matrix $\mathbf{P} \in \mathbb{R}^{36 \times 36}$, and construct the symmetric matrix $\mathbf{M} = \mathbf{B}^\top \mathbf{B} \in \mathbb{R}^{6 \times 6}$. This matrix can be divided into rotational-translational coupling blocks:

\begin{equation}
    \mathbf{M} = \left[\begin{array}{cc}
    \mathbf{M}_{r r} & \mathbf{M}_{r p} \\
    \mathbf{M}_{p r} & \mathbf{M}_{p p}
    \end{array}\right]_{6\times 6},
    \label{eq_20_degenerate_p}
\end{equation}
where $\mathbf{M}_{r r} \in \mathbb{R}^{3 \times 3}$ and $\mathbf{M}_{p p} \in \mathbb{R}^{3 \times 3}$ contain the covariance information for rotational and translational variables, respectively. $\mathbf{M}_{p p}$ and $\mathbf{M}_{r r}$ describe their coupling relationships. We then perform eigenvalue decomposition on the rotational and translational components:
\begin{equation}
    \mathbf{M}_{p p} = \mathbf{V}_p \boldsymbol{\Sigma}_p \mathbf{V}_p^{\top}, \quad \mathbf{M}_{r r} = \mathbf{V}_r \boldsymbol{\Sigma}_r \mathbf{V}_r^{\top},
    \label{eq_21_degenerate_decompose}
\end{equation}
where $\mathbf{V}_p = [\mathbf{v}_{p_1}, \mathbf{v}_{p_2}, \mathbf{v}_{p_3}]$ and $\mathbf{V}_r = [\mathbf{v}_{r_1}, \mathbf{v}_{r_2}, \mathbf{v}_{r_3}]$ are orthogonal eigenvector matrices, and $\boldsymbol{\Sigma}_p = \operatorname{diag}(\sigma_{p_1}, \sigma_{p_2}, \sigma_{p_3})$ and $\boldsymbol{\Sigma}_r = \operatorname{diag}(\sigma_{r_1}, \sigma_{r_2}, \sigma_{r_3})$ are normalized eigenvalue diagonal matrices, satisfying $\sigma_{p_1} \geq \sigma_{p_2} \geq \sigma_{p_3}$ and $\sigma_{r_1} \geq \sigma_{r_2} \geq \sigma_{r_3}$, respectively. The SLAM system is considered to have degraded in the directions $\mathbf{v}_{p_{\max}}$ and $\mathbf{v}_{r_{\max}}$ when the following conditions are met:
\begin{equation}
    \begin{cases}
    \sigma_{p_{\max}} \geq D_{p}^{\text{thre}} \\[0.5em]
    \sigma_{r_{\max}} \geq D_{r}^{\text{thre}}
    \end{cases},
    \label{eq_22_condition}
\end{equation}
where $\sigma_{p_{\max}} \triangleq \sigma_{p_1}$ and $\sigma_{r_{\max}} \triangleq \sigma_{r_1}$ are the maximum eigenvalues, and $D_{p}^{\text{thre}}$ and $D_{r}^{\text{thre}}$ are the degradation thresholds for translation and rotation, respectively.

\subsubsection{Adaptive Motion Mode Switching}

We propose an adaptive motion mode switching mechanism based on degradation detection. First, we determine whether the robot is within the UWB measurement range to decide whether UWB measurements should be incorporated. Next, using the eigenvalues and eigenvectors obtained from degradation detection, we evaluate the degree and direction of degradation to determine whether wheel odometer constraints should be introduced to enhance the system's robustness.

As illustrated in Fig. \ref{Fig4-Dengerate_detection}-(d), the maximum measurement range of the UWB positioning system forms a three-dimensional sphere centered at $\mathbf{P}_{0}$, with a maximum measurement radius $r_{\max}$. This region is defined as:
\begin{equation}
    \mathcal{R}_U = \left\{ \mathbf{P} \in \mathbb{R}^3 \mid \|\mathbf{P} - \mathbf{P}_0\| \leq r_{\max} \right\},
    \label{eq_23_uwb_area}
\end{equation}
where $\mathbf{P}_{end}$ is the farthest boundary point measured by the UWB along the tunnel, and $\mathbf{P}_{robot}$ denotes the robot's position in the global coordinate system $G$.

Our degradation detection and adaptive motion mode switching mechanism operates as follows:

\textbf{STEP1 - LIU Mode:}
As shown in Fig. \ref{Fig4-Dengerate_detection}-(a), when $\mathbf{P}_{robot} \in \mathcal{R}_U$ is satisfied, the SLAM system propagates the state in LIU mode. UWB positioning measurements $\mathbf{z}_{U_p}$ are used to align the system with the global coordinate system $G$, while UWB distance measurements $\mathbf{z}_{U_r}$ provide precise localization.

\textbf{STEP2 - LIO Mode:}
As shown in Fig. \ref{Fig4-Dengerate_detection}-(b), when the robot moves beyond the UWB measurement range (i.e., $\mathbf{P}_{robot} \notin \mathcal{R}_U$), the UWB distance constraint $\mathbf{r}_{U_r}$ and UWB position constraint $\mathbf{r}_{U_p}$ are discarded, and the system switches to LIO mode.

\textbf{STEP3 - LIW Mode:}
As shown in Fig. \ref{Fig4-Dengerate_detection}-(c), when the robot moves outside the UWB measurement range, the system's degradation degree and direction are calculated using the eigenvalues and eigenvectors from degradation detection. If the following conditions are met:
\begin{equation}
    \begin{cases}
    \mathbf{P}_{robot} \notin \mathcal{R}_U  \\[0.5em]
    \sigma_{p_{\max}} \geq D_{p_{thre}} \\
    \sigma_{r_{\max}} \geq D_{r_{thre}}
    \end{cases}, 
    \label{eq_24_uwb_area_judge}
\end{equation}
the system is considered to have degenerated. In this case, we fuse wheel odometer measurements into the SLAM system to assist the LIO system in state estimation within degraded tunnel environments without UWB measurements.

\section{Experiment}\label{chap6-experiment}
\subsection{Experiment Setup}

Field experiments were conducted in a highly degraded underground coal mine tunnel environment using the CUMT\_5 mobile robot, equipped with a Robosens 32 LiDAR, Xsens-G710 IMU, p440 UWB module, and wheel odometer. To evaluate the algorithm's accuracy, the robot's trajectory ground truth was obtained using a total station, serving as a benchmark for precision assessment. The system was configured with an Intel i7 CPU, 32GB DDR4 RAM, NVIDIA GTX 1050Ti GPU, and a 512GB SSD. The performance of our proposed CM-LIUW-Odometry was evaluated in four modes—LIO+UWB+Wheel (Ours), LIO+UWB (Ours w/o wheel), LIO+Wheel (Ours w/o uwb), and LIO (Ours w/o wheel \& uwb)—and compared against four state-of-the-art algorithms: FAST-LIO2, DLIO, IG-LIO, and LIO-SAM.

\subsection{Realworld Experiment}

\begin{figure}[t]
    \centering
    \includegraphics[width=1\columnwidth]{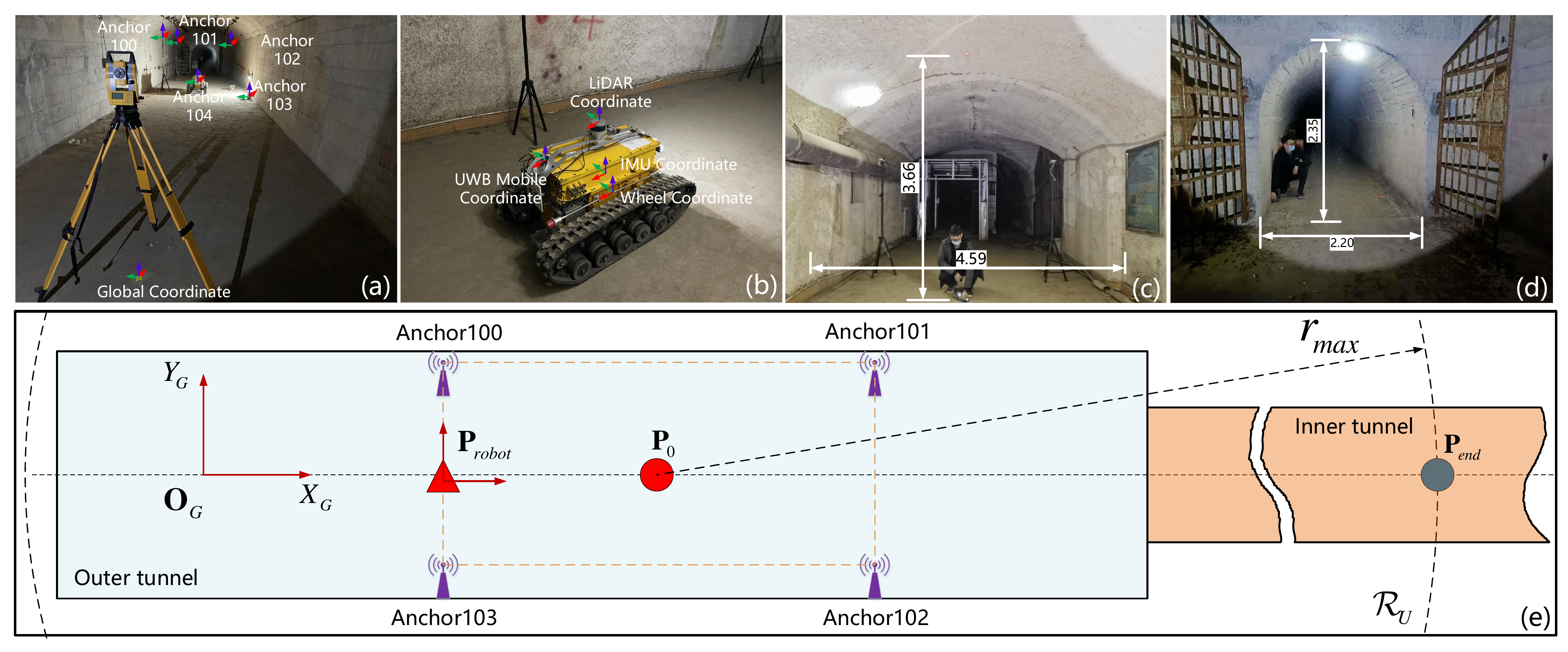}
    \caption{Field experiment environment in the underground coal mine tunnel. (a) Shows the deployment locations of UWB, CUMT\_5 robot, and total station in the underground coal mine tunnel. (b) Shows the sensor layout of LiDAR, UWB, IMU, and wheel odometer on CUMT\_5. (c) and (d) Show the environment and texture details of the outer tunnel and inner tunnel, respectively. (e) Shows the locations of UWB, CUMT\_5 robot, and total station in the underground coal mine tunnel, as well as the maximum measurement range $\mathcal{R}_U$ of the UWB positioning system.}
    \label{Fig5-field-exp-setup}
    \vspace{0cm}
\end{figure}

This section evaluates the trajectory accuracy of the proposed method in underground coal mine tunnels and the effectiveness of its degradation detection and adaptive switching mechanism. The experiments were conducted on the CUMT\_5 mobile robot platform in an extremely degraded tunnel environment. As shown in Fig. \ref{Fig5-field-exp-setup}-(c) and Fig. \ref{Fig5-field-exp-setup}-(d), the outer tunnel is relatively short and features rich textures, while the inner tunnel consists of long, straight sections with low-texture features. This scenario presents dual challenges for LiDAR SLAM: (1) long, straight tunnels reduce the detectability of geometric features, and (2) highly repetitive ground and wall features can lead to LiDAR degradation issues.

Fig. \ref{Fig5-field-exp-setup}-(a) and Fig. \ref{Fig5-field-exp-setup}-(e) illustrate the spatial layout of the field experiment. The total station center $\mathbf{O}_G$ is set as the global coordinate origin, with four UWB anchor nodes placed on both sides of the tunnel at coordinates Anchor100 (11.376, 1.694, 2.249), Anchor101 (16.678, 1.769, 2.247), Anchor102 (16.550, -1.453, 2.224), and Anchor103 (11.510, -1.532, 0.115). The robot's initial position is $\mathbf{P}_{robot}$ (11.490, -0.019, 0.971), and the center of the UWB positioning system is located at $\mathbf{P}_{0}$ (13.963, 0, 2.249). The maximum measurement range $\mathcal{R}_U$ of the UWB positioning system forms a three-dimensional sphere centered at $\mathbf{P}_{0}$ with a maximum measurement radius $r_{\max}$. The farthest effective point along the tunnel direction that the UWB can measure is $\mathbf{P}_{end}$ (34.963, 0, 2.249). UWB observation updates cease when $\mathbf{P}_{robot} \notin \mathcal{R}_U$. Fig. \ref{Fig5-field-exp-setup}-(b) shows the layout of all sensors on the CUMT\_5 robot, which moves at a constant speed of 0.3 m/s with a maximum angular velocity of 0.2 rad/s.  

\subsubsection{Trajectory Accuracy Analysis}
\begin{figure*} 
    \centering
    \includegraphics[width=1\textwidth]{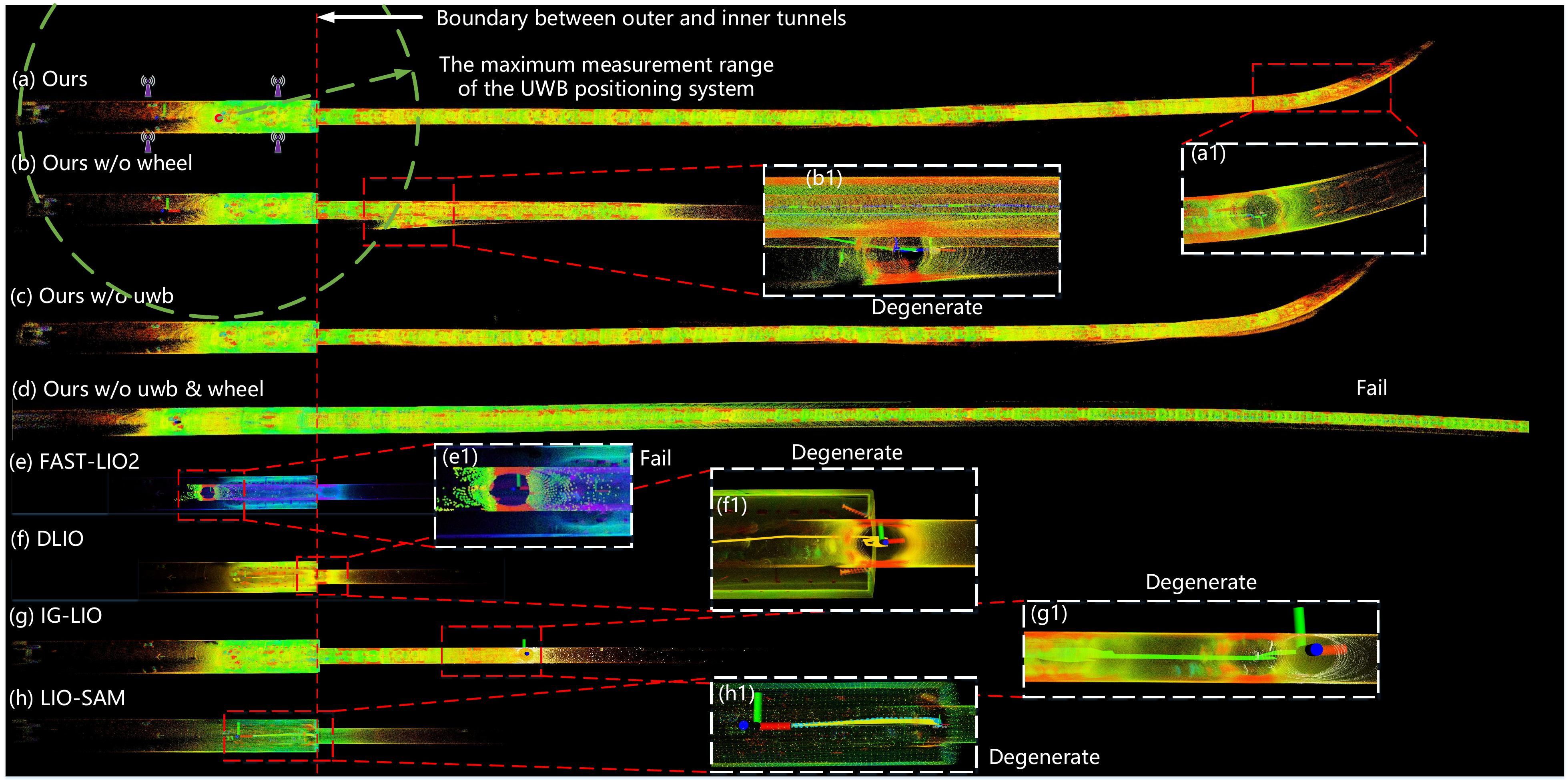}
    \caption{Mapping performance of (a) Ours, (b) Ours w/o wheel, (c) Ours w/o uwb, (d) Ours w/o wheel \& uwb, (e) FAST-LIO2, (f) DLIO, (g) IG-LIO, and (h) LIO-SAM in an underground coal mine environment. Significant degradation in the inner tunnel is observed for Ours w/o wheel, DLIO, IG-LIO, and LIO-SAM. Ours w/o uwb \& wheel and FAST-LIO2 fail entirely, while Ours and Ours w/o wheel successfully navigate to the tunnel endpoint.}
    \label{Fig6-field-exp-mapping}
\end{figure*}

\begin{figure}[htpb]
    \centering
    \includegraphics[width=1\columnwidth]{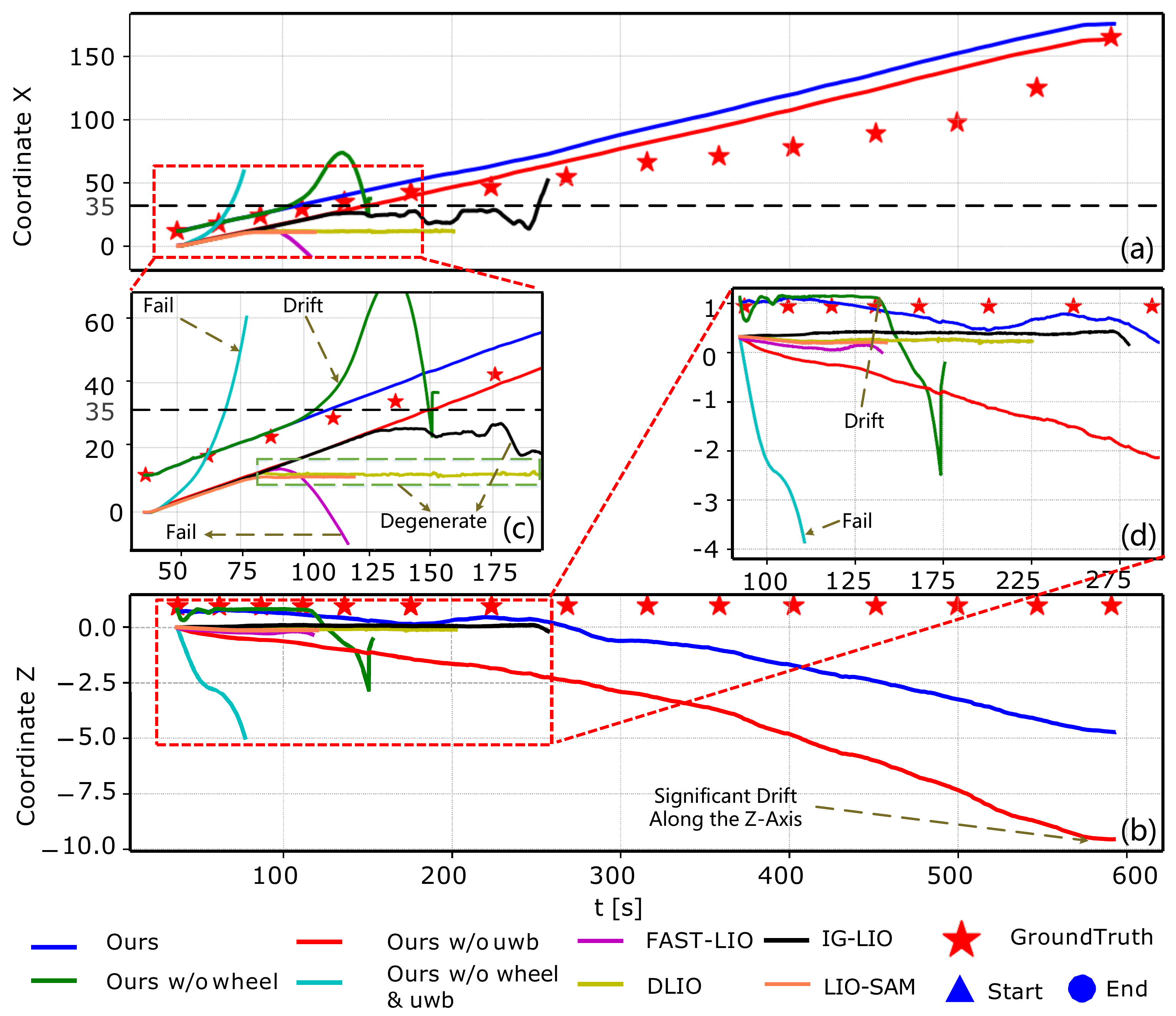}
    \caption{Time-dependent X- and Z-axis trajectory components. (a, b) Full trajectories for Ours, Ours w/o wheel, Ours w/o UWB, Ours w/o wheel \& UWB, FAST-LIO2, DLIO, IG-LIO, LIO-SAM, and Ground Truth. (c, d) Zoomed-in views highlighting key trajectory segments.}
    \label{Fig7-field-exp-traj}
    \vspace{0cm}
\end{figure}

Fig. \ref{Fig6-field-exp-mapping} and Fig. \ref{Fig7-field-exp-traj} present the mapping results and trajectory evaluation for CM-LIUW-Odometry and mainstream SLAM algorithms in a real underground coal mine tunnel. (Note: We define "drift" and "degradation" as follows: Drift refers to a continuous, unidirectional offset of the odometry during the robot's movement, while degradation refers to small, back-and-forth oscillations of the odometry.) Due to the lack of GPS in the underground coal mine tunnel, we evaluate the trajectory using 15 ground truth points, $\mathbf{p}_{gt}^i$, collected via a total station, and the estimated trajectory points, $\mathbf{p}_{es}^i$, from the algorithms. The total localization error is defined as $\mathrm{TotalErr}=\sum_{i=0}^{N-1} ||\mathbf{p}_{gt}^i-\mathbf{p}_{es}^i||$ and the average localization error as $\mathrm{AvgErr}=\sum_{i=0}^{N-1} \frac{||\mathbf{p}_{es}^i-\mathbf{p}_{gt}^i||}{N}$.

All comparison algorithms exhibited significant anomalies. FAST-LIO2 and Ours w/o uwb \& wheel both experienced drift in the negative/positive direction along the X-axis, ultimately leading to system failure (see Fig. \ref{Fig6-field-exp-mapping}-(e) and Fig. \ref{Fig6-field-exp-mapping}-(d)). DLIO and LIO-SAM experienced degradation at the junction between the outer and inner tunnels (see Fig. \ref{Fig6-field-exp-mapping}-(f) and Fig. \ref{Fig6-field-exp-mapping}-(h)). IG-LIO degraded and drifted after traveling 25 meters (see Fig. \ref{Fig6-field-exp-mapping}-(g)). Ours w/o wheel, which incorporates the UWB constraint, ensures precise pose estimation and alignment with the global coordinate system $G$ within $\mathcal{R}_U$, but drifts 25 meters along the X direction after exiting the area, entering a degraded state (see Fig. \ref{Fig7-field-exp-traj}-(c)). Ours w/o uwb, which uses wheel odometer, steadily reaches the end of the tunnel, but the accumulated error along the Z-axis is as high as 9.6 meters (see Fig. \ref{Fig7-field-exp-traj}-(b)). The complete system, Ours, aligns with the global coordinate system $G$ within $\mathcal{R}_U$ and effectively suppresses drift along the Z-axis, maintaining positioning stability via wheel odometer after exiting $\mathcal{R}_U$.

Table \ref{tbl2:localization_error} shows that Ours achieves a $\mathrm{TotalErr}$ of 73.789 and an $\mathrm{AvgErr}$ of 4.919, significantly outperforming the comparison algorithms. Ours w/o wheel and Ours w/o uwb have $\mathrm{TotalErr}$ values of 535.787 and 129.351, respectively, while DLIO and LIO-SAM, due to severe degradation, exhibit $\mathrm{TotalErr}$ and $\mathrm{AvgErr}$ values exceeding 700 and 50. This experiment validates the effectiveness of the multi-sensor fusion strategy, demonstrating that the removal of any sensor significantly degrades system performance.


\begin{table}[htpb]
    \centering
    \caption{Analysis of total localization error and average localization error.}\label{tbl2:localization_error}
    \begin{threeparttable}
        \begin{tabularx}{\linewidth}{@{}l *{6}{>{\centering\arraybackslash}X}@{}}
            \toprule
             & \rotatebox{0}{\makecell{Ours}} 
             & \rotatebox{0}{\makecell{Ours w/o \\ wheel}} 
             & \rotatebox{0}{\makecell{Ours w/o \\ uwb}} 
             & \rotatebox{0}{\makecell{DLIO}} 
             & \rotatebox{0}{\makecell{IG-LIO}}
             & \rotatebox{0}{\makecell{LIO-\\SAM}} \\
            \midrule
            $\mathrm{TotalErr}$ &   \textbf{73.789} & 535.787 &  \underline{129.35} & 790.211  & 582.762 & 800.586 \\
            $\mathrm{AvgErr}$ &  \textbf{4.919}	 & 35.719	& \underline{8.623} & 52.681 & 38.851 & 53.372  \\
            \bottomrule
        \end{tabularx}
        \begin{tablenotes}
            \item \textbf{Bold} indicates the best accuracy, \underline{underline} indicates the second best.
        \end{tablenotes}
    \end{threeparttable}
\end{table}

\subsubsection{Analysis of Degradation Detection and Adaptive Motion Mode Switching}  

\begin{figure}[h]
    \centering
    \includegraphics[width=1\columnwidth]{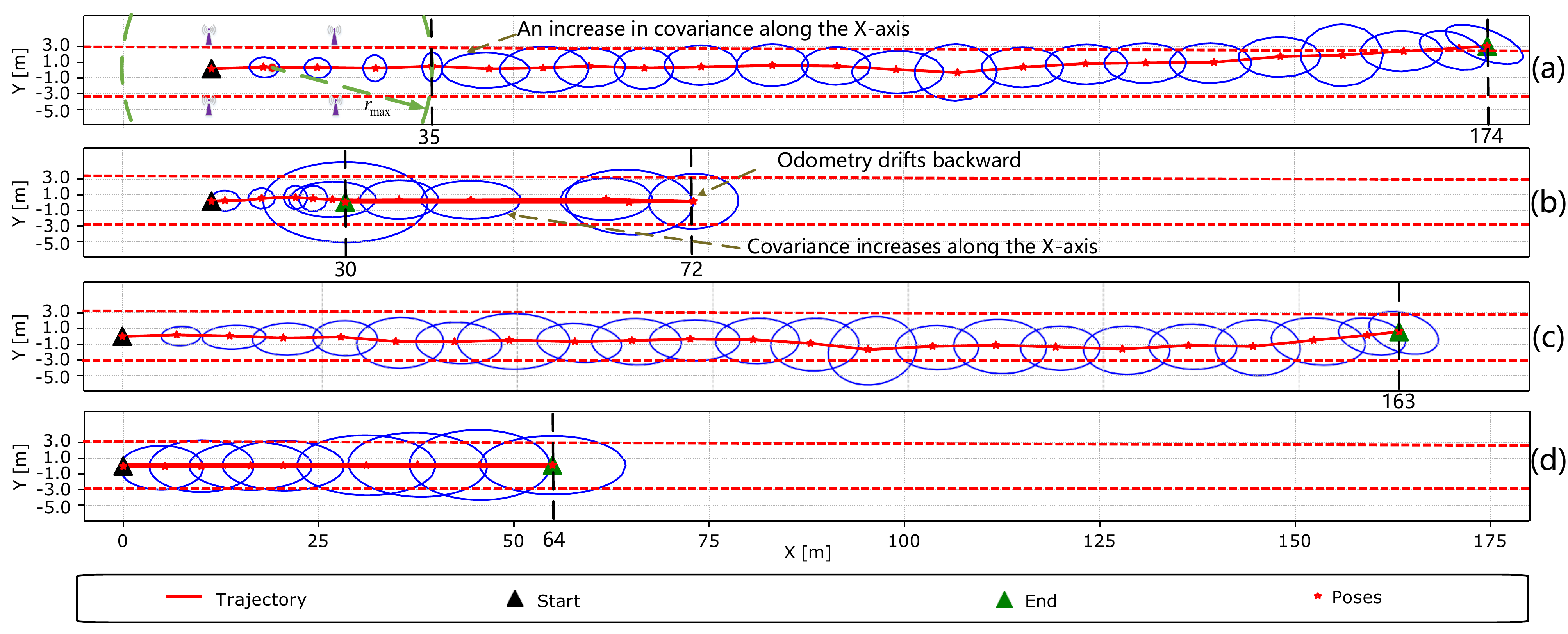}
    \caption{Covariance ellipsoids. (a)-(d) respectively show the covariance ellipsoids for Ours, Ours w/o wheel, Ours w/o uwb, and Ours w/o wheel \& uwb in a real underground coal mine tunnel. The red dashed lines represent the boundary of the Y-axis covariance component, with values of +3.0 and -3.0.}
    \label{Fig8-field-exp-cov}
    \vspace{0cm}
\end{figure}

\begin{figure}[t]
    \centering
    \includegraphics[width=1\columnwidth]{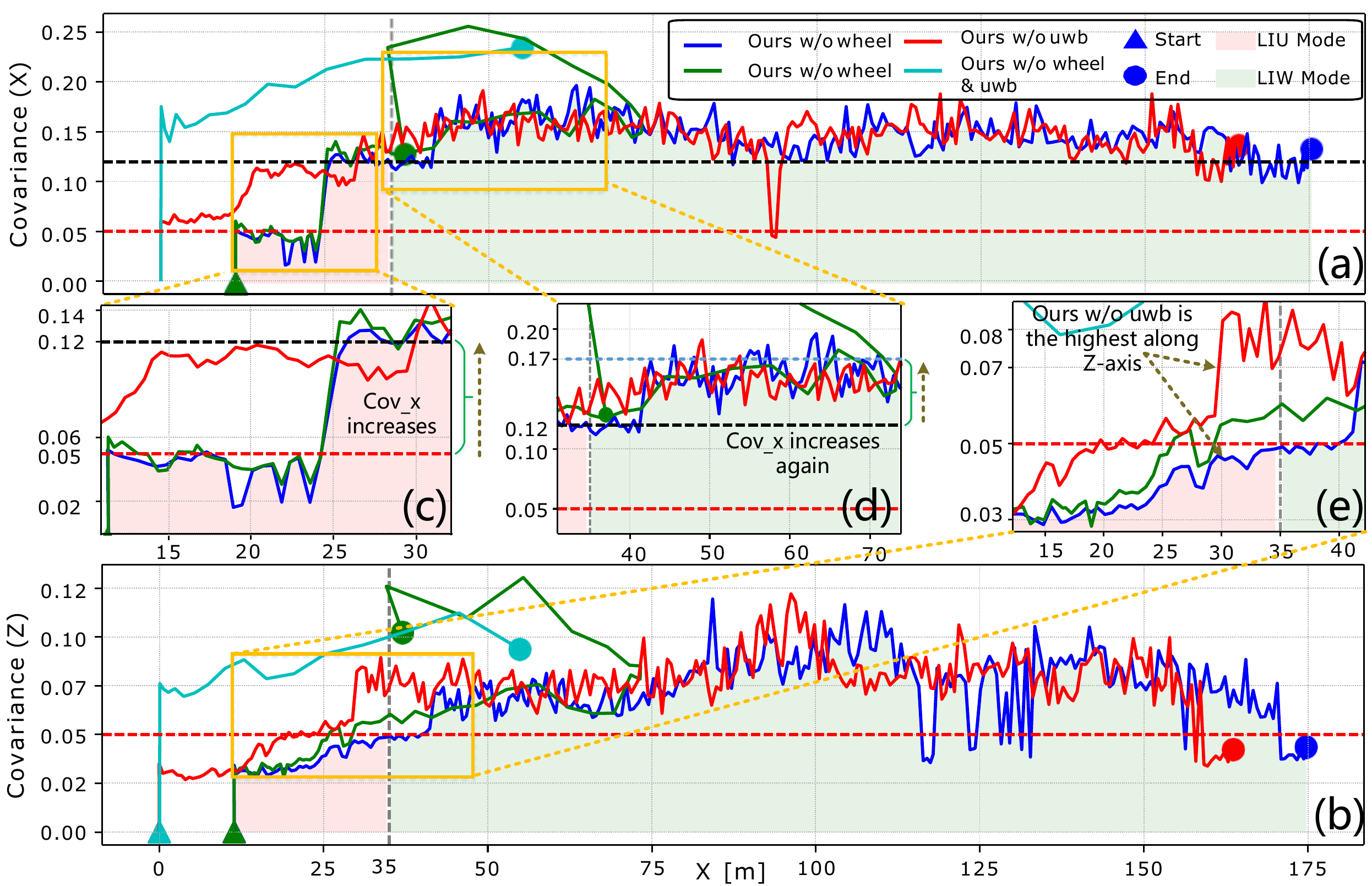}
    \caption{Covariance analysis in the X and Z directions. (a) and (b) respectively show the X and Z direction covariance for Ours, Ours w/o wheel, Ours w/o uwb, and Ours w/o wheel \& uwb in a real underground coal mine tunnel. (c) and (d) show zoomed-in views of the X direction covariance in the 11-30m and 35-75m ranges along the tunnel. (e) shows a zoomed-in view of the Z direction covariance in the 11-40m range along the tunnel.}
    \label{Fig9-field-exp-cov-analysis}
    \vspace{0cm}
\end{figure}

Fig. \ref{Fig8-field-exp-cov} and Fig. \ref{Fig9-field-exp-cov-analysis} present the covariance analysis for four configurations of CM-LIUW-Odometry. Given that only translational degradation along the X-axis is observed in the tunnel scenario, rotational degradation is not considered. We set $D_{p}^{thre} = 0.05$ as the threshold for degradation detection. (Note: In Fig. \ref{Fig8-field-exp-cov}, the covariance components along the X and Y axes are magnified by a factor of 100 to visualize covariance variations during motion mode switching, while the covariance values in Fig. \ref{Fig9-field-exp-cov-analysis} remain unaltered.)

The complete system (Ours) demonstrates the minimal covariance in the X/Z directions within $\mathcal{R}_U$ (see Fig. \ref{Fig8-field-exp-cov}-(a), Fig. \ref{Fig9-field-exp-cov-analysis}-(a) and Fig. \ref{Fig9-field-exp-cov-analysis}-(c)). Upon exiting $\mathcal{R}_U$, it switches to the LIW mode using wheel odometer constraints to maintain stability. In contrast, Ours w/o wheel suffers from X/Z covariance surges outside $\mathcal{R}_U$, leading to drift (Fig. \ref{Fig8-field-exp-cov}-(b) and Fig. \ref{Fig9-field-exp-cov-analysis}-(a)). Ours w/o uwb, lacking precise position constraints from UWB, exhibits Z-axis covariance exceeding $D_{p}^{thre}$ in most regions, resulting in significant cumulative Z-axis errors (Fig. \ref{Fig9-field-exp-cov-analysis}-(e)). Ours w/o uwb \& wheel starts with initial covariance exceeding $D_{p}^{thre}$, ultimately causing system failure (Fig. \ref{Fig9-field-exp-cov-analysis}-(a)). These experiments validate that multimodal constraints synergistically suppress covariance growth and enhance the localization robustness of the SLAM system in degraded environments.

\section{Conclusion}\label{chap7-conclusion}  

This paper introduces CM-LIUW-Odometry, a multimodal SLAM framework integrating LiDAR, IMU, UWB, and wheel odometer, specifically designed for large-scale, complex, and degraded underground coal mine environments. The algorithm employs a tightly coupled approach to fuse LiDAR-inertial odometry with UWB absolute positioning constraints, effectively creating a GPS-like positioning system for underground environments. Furthermore, the wheel odometer is tightly integrated into the fusion framework to address SLAM system degradation in areas outside the UWB measurement range. A novel motion mode switching mechanism is proposed, enabling adaptive transitions between modes based on the detected level of environmental degradation. The proposed method is extensively evaluated and compared with state-of-the-art approaches in real-world scenarios, demonstrating superior robustness and reliability in challenging underground coal mine environments.

\bibliographystyle{IEEEtranBST/IEEEtranS}
\bibliography{IEEEtranBST/ref}

\end{document}